\documentclass[fleqn,10pt]{wlscirep}
\usepackage[utf8]{inputenc}
\usepackage[T1]{fontenc}
\usepackage{tabularx}
\usepackage{amsmath}
\usepackage{multirow}
\usepackage{xcolor}
\usepackage{setspace}
\usepackage{csquotes}
\usepackage[backend=biber, style=numeric-comp, sorting=none, maxnames=3, minnames=3, giveninits=true, doi=false, isbn=false, url=true, eprint=false]{biblatex}
\DeclareNameAlias{sortname}{family-given}
\DeclareNameAlias{default}{family-given}
\renewrobustcmd*{\bibinitdelim}{} 

\DeclareFieldFormat[article, inbook, incollection, inproceedings, misc, thesis]{title}{#1}
\DeclareFieldFormat[book, inbook, incollection, inproceedings, misc, thesis]{title}{\mkbibemph{#1}}

\renewbibmacro{in:}{}

\DeclareFieldFormat[article]{volume}{#1}
\DeclareFieldFormat[article]{number}{\mkbibparens{#1}}

\DeclareFieldFormat[article]{pages}{#1}

\renewbibmacro*{volume+number+eid}{%
  \setunit{\addsemicolon}%
  \printfield{volume}%
  \setunit*{\addcolon}%
  \printfield{pages}}

\renewbibmacro*{journal+issuetitle}{%
  \usebibmacro{journal}%
  \setunit*{\space}%
  \printfield{year}%
  \setunit*{\addsemicolon}%
  \printfield{volume}%
  \newunit}

\renewbibmacro*{issue+date}{%
  \setunit*{\addcomma\space}%
  \printtext[parens]{\printfield{year}}%
  \newunit}

\renewbibmacro*{note+pages}{%
  \setunit{\addcolon}
  \printfield{pages}%
  \newunit}

\DeclareFieldFormat[misc]{title}{#1} 
\DeclareFieldFormat[misc]{howpublished}{\mkbibemph{#1}} 
\DeclareFieldFormat[misc]{url}{\url{#1}} 
\DeclareFieldFormat[misc]{note}{#1} 

\definecolor{customblue}{RGB}{56, 84, 146}
\definecolor{customgrey}{RGB}{89, 89, 89}
\definecolor{customgreen}{RGB}{94, 129, 63}
\definecolor{customyellow}{RGB}{184, 146, 48}

\title{A dataset and benchmark for hospital course summarization with adapted large language models}

\author[1, 2,*]{Asad Aali, MS}
\author[3, 4]{Dave Van Veen, PhD}
\author[2]{Yamin Ishraq Arefeen, PhD}
\author[5]{Jason Hom, MD}
\author[5, 6]{Christian Bluethgen, MS, MD}
\author[3, 7]{Eduardo Pontes Reis, MD}
\author[1, 3]{Sergios Gatidis, MD}
\author[8]{Namuun Clifford, MSN, FNP}
\author[9]{Joseph Daws, PhD}
\author[9]{Arash S. Tehrani, PhD}
\author[10]{Jangwon Kim, PhD}
\author[1, 3, 11]{Akshay S. Chaudhari, PhD}

\affil[1]{Department of Radiology, Stanford University, Stanford, CA, USA}
\affil[2]{Department of Electrical and Computer Engineering, The University of Texas at Austin, Austin, TX, USA}
\affil[3]{Center for Artificial Intelligence in Medicine and Imaging, Stanford University, Palo Alto, CA, USA}
\affil[4]{Department of Electrical Engineering, Stanford University, Stanford, CA, USA}
\affil[5]{Department of Medicine, Stanford, CA, USA}
\affil[6]{University Hospital Zurich, Zurich, Switzerland}
\affil[7]{Albert Einstein Israelite Hospital, São Paulo, Brazil}
\affil[8]{School of Nursing, The University of Texas at Austin, Austin, TX, USA}
\affil[9]{One Medical, San Francisco, CA, USA}
\affil[10]{Amazon, Seattle, WA, USA}
\affil[11]{Department of Biomedical Data Science, Stanford University, Stanford, CA, USA\newline}
\affil[*]{\textbf{Corresponding Author:}\par 
\textbf{Name:} Asad Aali\par
\textbf{Institute:} Stanford University\par
\textbf{Address:} 1701 Page Mill Rd,
Palo Alto, CA, 94304, USA\par 
\textbf{Email:} asadaali@stanford.edu}

\begin{abstract}
\textbf{Objective:} Brief hospital course (BHC) summaries are clinical documents that summarize a patient's hospital stay. While large language models (LLMs) depict remarkable capabilities in automating real-world tasks, their capabilities for healthcare applications such as synthesizing BHCs from clinical notes have not been shown. We introduce a novel pre-processed dataset, the MIMIC-IV-BHC, encapsulating clinical note and brief hospital course (BHC) pairs to adapt LLMs for BHC synthesis. Furthermore, we introduce a benchmark of the summarization performance of two general-purpose LLMs and three healthcare-adapted LLMs.\\

\textbf{Materials and Methods:} Using clinical notes as input, we apply prompting-based (using in-context learning) and fine-tuning-based adaptation strategies to three open-source LLMs (Clinical-T5-Large, Llama2-13B, FLAN-UL2) and two proprietary LLMs (GPT-3.5, GPT-4). We evaluate these LLMs across multiple context-length inputs using natural language similarity metrics. We further conduct a clinical study with five clinicians, comparing clinician-written and LLM-generated BHCs across 30 samples, focusing on their potential to enhance clinical decision-making through improved summary quality. We compare reader preferences for the original and LLM-generated summary using Wilcoxon Signed-Rank tests. We further request optional qualitative feedback from clinicians to gain deeper insights into their preferences, and we present the frequency of common themes arising from these comments.\\

\textbf{Results:} The Llama2-13B fine-tuned LLM outperforms other domain-adapted models given quantitative evaluation metrics of BLEU and BERT-Score. GPT-4 with in-context learning shows more robustness to increasing context lengths of clinical note inputs than fine-tuned Llama2-13B. Despite comparable quantitative metrics, the reader study depicts a significant preference for summaries generated by GPT-4 with in-context learning compared to both Llama2-13B fine-tuned summaries and the original summaries ($p < 0.001$), highlighting the need for qualitative clinical evaluation. \\

\textbf{Discussion and Conclusion:} We release a foundational clinically relevant dataset, the MIMIC-IV-BHC, and present an open-source benchmark of LLM performance in BHC synthesis from clinical notes. We observe high-quality summarization performance for both in-context proprietary and fine-tuned open-source LLMs using both quantitative metrics and a qualitative clinical reader study. Our research effectively integrates elements from the data assimilation pipeline: our methods use (1) \textit{clinical data sources} to integrate (2) \textit{data translation} and (3) \textit{knowledge creation}, while our evaluation strategy paves the way for (4) \textit{deployment}.\\


\textbf{Word Count:} $3,905$
\end{abstract}

\begin{document}
\doublespacing
\maketitle
\clearpage

\section*{OBJECTIVE}

Clinicians spend significant time on clinical documentation \cite{moy2021measurement, chaiyachati2019assessment, mamykina2016residents}, a crucial component of data assimilation \cite{albers2020assimilation} beginning with \textit{data sources} such as EHRs and progressing through \textit{data translation} and \textit{knowledge creation} where these records can be processed for clinically-relevant machine learning tasks. The discharge summary (clinical note) is a key document clinicians refer to in important scenarios, including when a discharged patient presents to the clinic for post-hospitalization follow-up. The brief hospital course (BHC) fulfills an important role within the discharge summary, particularly for long and complex hospitalizations. Clinicians who are pressed for time may choose to focus on the BHC summary instead of the entire discharge summary. Synthesizing a BHC given a clinical note requires substantial clinician time and expertise, with errors possibly causing patient harm \cite{clough2024transforming}. Automating BHC generation is important since discharge summaries can lack essential information, contain incorrect information, and may be incomplete before follow-up appointments \cite{kripalani2007deficits}. The success of large language models (LLMs) has provided a promising avenue for this clinically important task of BHC synthesis. Having LLMs generate the first draft of a BHC may lead to fewer errors by reducing cognitive load and fatigue-related mistakes, produce higher-quality summaries by consistently incorporating all relevant clinical details, and substantially reduce clinician time spent on documentation by automating the initial drafting process \cite{patel2023chatgpt, singhal2023large, van2024clinical}. However, while LLMs can produce high-quality drafts, clinical use should consider automation bias, where these drafts could be accepted as final without a thorough review, requiring interfaces designed to promote critical assessment and evaluation \cite{warraich2024fda}.

\section*{BACKGROUND AND SIGNIFICANCE}
Recent advancements in natural language processing (NLP) have been characterized by transformer-based language models \cite{NIPS2017_3f5ee243}, such as Bidirectional Encoder Representations from Transformers (BERT) \cite{kenton2019bert} and Unidirectional Generative Pre-trained Transformer 2 (GPT-2) \cite{radford2019language}. The evolution of transformer models involved pretraining on large corpora of text and subsequent fine-tuning on domain-specific data, as demonstrated by GPT-3 \cite{NEURIPS2020_1457c0d6}, PaLM \cite{chowdhery2023palm}, and T5 \cite{raffel2020exploring}. After pre-training, these LLMs are often exposed to supervised fine-tuning given input-output pairs to improve their ability to follow instructions (instruction turning) \cite{zhang2023instruction}. Following instruction tuning, reinforcement learning with human feedback (RLHF) is employed to align the LLM with human preferences \cite{zhang2023instruction, wang2023pre}. A recent work presented a multi-document brief hospital course dataset to promote the advancement of hospital visit summarization \cite{adams2021s}.

Automating documentation tasks such as BHC synthesis can potentially alleviate clinician workload and improve documentation quality. While previous works have explored deep learning-based automation for hospital course summarization \cite{searle2023discharge, hartman2023method, jung2024enhancing}, no open-source benchmark exists to evaluate state-of-the-art (SOTA) LLM performance on BHC synthesis. Developing a benchmark and standardized framework for evaluating model effectiveness can help advance NLP applications in reducing clinical documentation burdens. Our study addresses this gap by introducing a foundational task-specific dataset, MIMIC-IV-BHC \cite{aali2024mimic}, which is tailored specifically for BHC summarization. By benchmarking SOTA LLMs on this dataset, we aim to facilitate the development of more efficient and accurate models for brief hospital course summarization. The significance of this work lies in its potential to reduce documentation errors, improve the quality of clinical summaries, and free up clinician time, which could directly contribute to enhancing patient care and safety.

\section*{MATERIALS AND METHODS}

\begin{table*}[t]
\caption{\textbf{a)} A sample from MIMIC-IV-BHC, our novel pre-processed dataset extracted from raw MIMIC-IV notes.}
\begin{tabular*}{\textwidth}{@{\extracolsep{\fill}}lll}
\toprule
\textbf{Input} & \textbf{Example} \\
\midrule
SEX & F \\
SERVICE & SURGERY \\
ALLERGIES & No Known Allergies \\
CHIEF COMPLAINT & Splenic laceration \\
MAJOR PROCEDURE & NONE \\
HISTORY OF PRESENT ILLNESS & s/p routine colonoscopy this morning with polypectomy (report not available) ... \\
PAST MEDICAL HISTORY & Mild asthma, hypothyroid \\
FAMILY HISTORY & Non-contributory \\
PHYSICAL EXAM & Gen: Awake and alert CV: RRR Lungs: CTAB Abd: Soft, nontender, nondistended \\
PERTINENT RESULTS & 03:45 PM BLOOD WBC-5.5 RBC-3.95 Hgb-14.1 ... \\
MEDICATIONS ON ADMISSION & 1. Levothyroxine Sodium 100 mcg PO DAILY 2. Flovent HFA (fluticasone) ... \\
DISCHARGE DISPOSITION & Home \\
DISCHARGE DIAGNOSIS & Splenic laceration \\
DISCHARGE CONDITION & Mental Status: Clear and coherent. Level of Consciousness: Alert and interactive ... \\
DISCHARGE INSTRUCTIONS & You were admitted to ... in the intensive care unit for monitoring after a ... \\ \\
\toprule
\textbf{Output} & \textbf{Example} \\
\midrule
BRIEF HOSPITAL COURSE & Ms. ... was admitted to ... on .... After getting a colonoscopy and polypectomy, she ... \\
\bottomrule
\label{table:dataset_sample}
\end{tabular*}
\\ \textbf{b)} Statistics for the MIMIC-IV-BHC subsets, binned across multiple context length ranges for adaptation tasks. \newline \\
\begin{tabular*}{\textwidth}{@{\extracolsep{\fill}}llll}
\toprule
\textbf{Context Range} & \textbf{Train/Test Split} & \textbf{Input Tokens} & \textbf{BHC Tokens} \\
\midrule
$0$ - $1,024$ & $2,000 / 100$ & $711 \pm 199$  & $104 \pm 43$\\
$1,024$ - $2,048$ & $2,000 / 100$ & $1,471 \pm 275$  & $148 \pm 36$\\
$2,048$ - $4,096$ & $2,000 / 100$ & $2,496 \pm 388$ & $225 \pm 55$\\
\end{tabular*}
\end{table*}

We utilize the MIMIC-IV-Note \cite{johnson2023mimic} dataset to curate a foundational task-specific dataset, the MIMIC-IV-BHC \cite{aali2024mimic}, tailored for the task of BHC summarization. Subsequently, we present an open-source benchmark of LLM performance on this dataset, across a variety of adaptation strategies.

\subsection*{Dataset}\label{sec4}
We utilize a raw clinical \textit{data source} \cite{albers2020assimilation} called MIMIC-IV-Note \cite{johnson2023mimic}, a compilation of 331,794 de-identified discharge summaries from 145,915 patients admitted to the Beth Israel Deaconess Medical Center. MIMIC-IV-Note \cite{johnson2023mimic} is a publicly available dataset reflective of real-world clinical notes. We pre-process MIMIC-IV \cite{johnson2023mimic} notes to create a labeled task-specific dataset, the MIMIC-IV-BHC \cite{aali2024mimic} (Table \ref{table:dataset_sample}), emphasizing the relationship between clinical notes and BHCs. Our open-source novel dataset is crucial to allow researchers to pursue further benchmarking studies. This study was institutional review board exempt because the MIMIC dataset does not meet the criteria for human subjects research described by our institution.

\begin{figure*}[t]
\begin{center}
\includegraphics[width=1\textwidth]{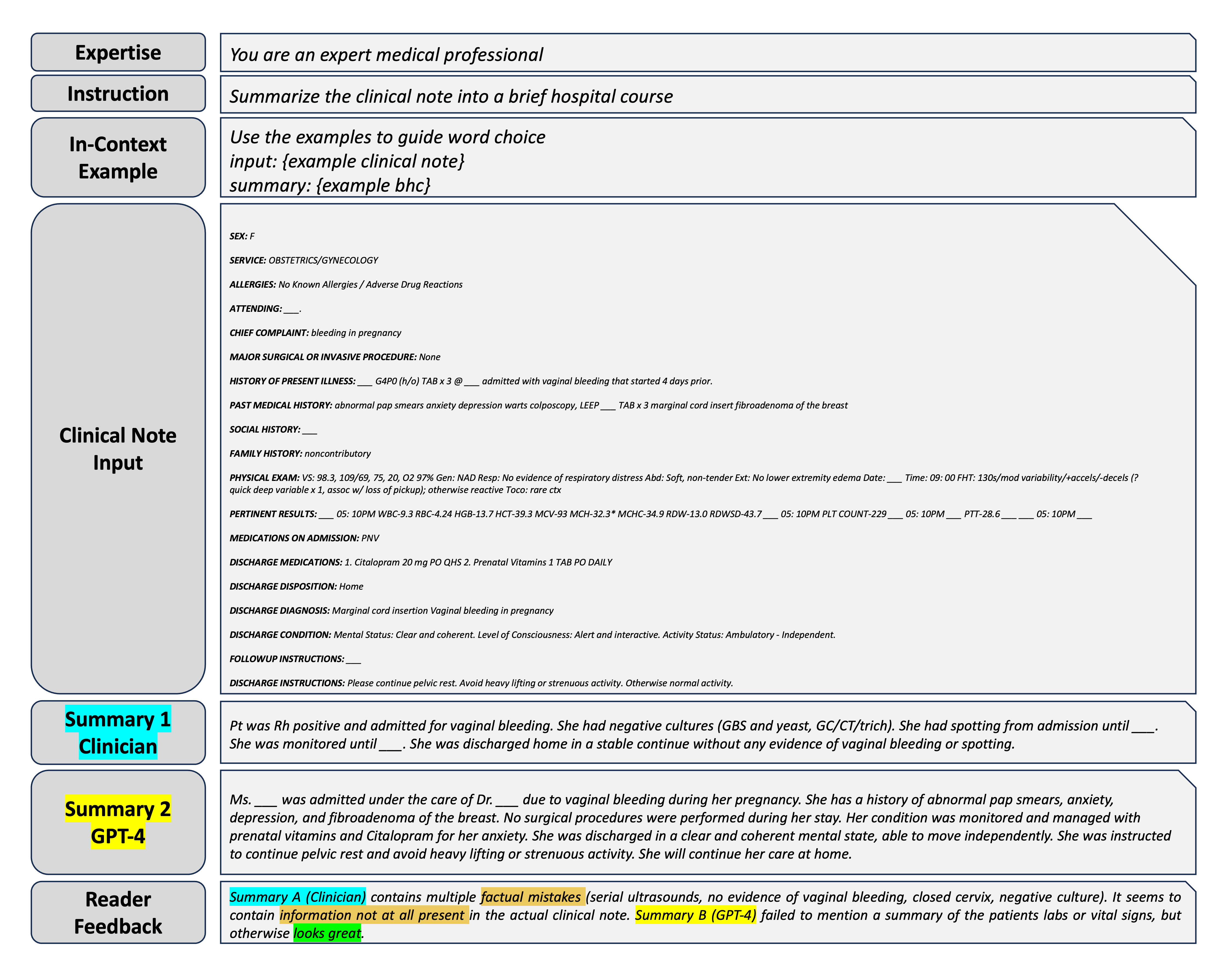}
\end{center}
\caption{A full-length clinical note with its respective clinician-written and LLM-generated BHC (with feedback). This note was sampled from the MIMIC-IV-BHC 0 to 1,024 context range subset. "Summary 1" is the actual BHC written by a clinician, and "Summary 2" is the BHC generated by GPT-4 adapted through in-context learning (ICL). The summaries were presented for feedback to the reader randomly, without specifying “clinician” or “GPT-4”.}
\label{fig:bhc_example}
\end{figure*}
\subsection*{MIMIC-IV-BHC Data Processing}

We perform \textit{data translation} \cite{albers2020assimilation}, a process of transforming raw, unstructured data into a standardized and structured format for downstream machine-learning tasks. This includes steps such as whitespace removal, section identification, tokenization, and other transformations that create a uniform structure for the MIMIC-IV-BHC dataset \cite{aali2024mimic}. Using regular expressions and natural language toolkit (NLTK) in Python, we apply: whitespace removal, uniform spacing, section identification, header standardization, extraneous symbol removal, and tokenization. We extract the BHC from each clinical note to create the labeled MIMIC-IV-BHC \cite{aali2024mimic} (Table \ref{table:dataset_sample}) dataset containing 270,033 clinical notes–BHC pairs with an average input token length of 2,267 $\pm$ 914 and an average output token length of 564 $\pm$ 410. We apply a uniform set of processing steps across all samples and manually review 100 randomly drawn samples from our dataset to validate the effectiveness of the pre-processing pipeline. The choice of 100 samples was based on a balance between confirming data quality and the resources available for manual review. Furthermore, the preprocessed dataset underwent an extensive review process before its publication on PhysioNet. Finally, our clinicians manually reviewed 30 clinical note-summary pairs from the dataset, without reporting any quality issues (Figure \ref{fig:bhc_example}).

The MIMIC-IV-BHC \cite{aali2024mimic} dataset is a function of the context length of the clinical notes. Recognizing that summarizing longer-length clinical notes is a challenging task in the context of text summarization, we further bin the dataset into three independent context length ranges for our adaptation tasks through random sampling as described in Table \ref{table:dataset_sample}b. For each subset in Table \ref{table:dataset_sample}b, we apply a minimum and maximum context length filter to the clinical notes and a minimum context length filter for the BHCs, helping remove note-BHC pairs with short texts, and ensuring the task’s focus on informative summarization. For fine-tuning tasks, we employ the 2,000 training samples within each context range from Table \ref{table:dataset_sample}b.

\subsection*{Large Language Models}\label{subsec2}

\begin{figure*}[t]
\begin{center}
\includegraphics[width=1\textwidth]{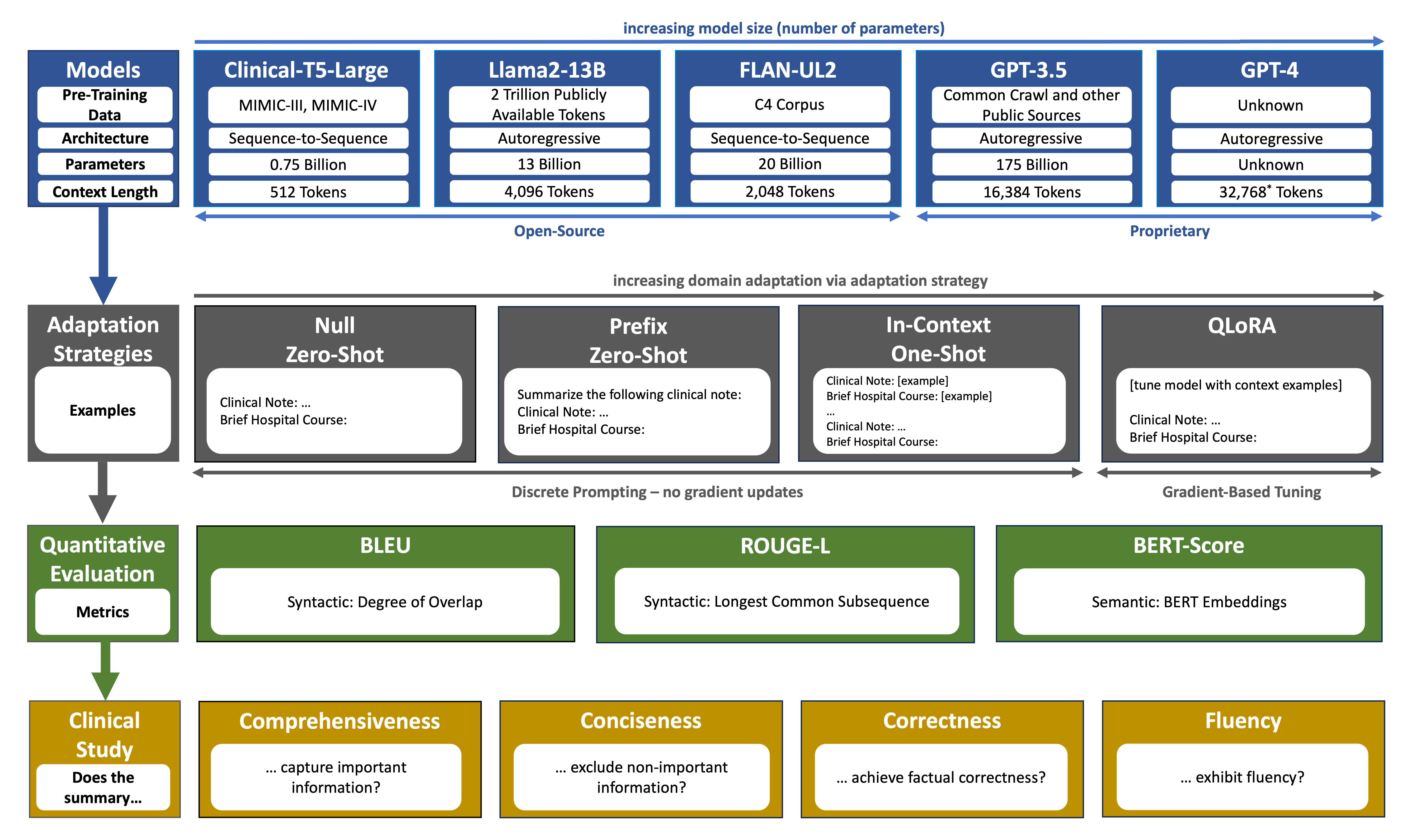}
\end{center}
\caption{Overall schematic of our study. We evaluate a variety of \textcolor{customblue}{\textbf{models}}, including open-source models containing up to 20 billion parameters, and larger-scale proprietary models. Each model is adapted to the summarization task using the \textcolor{customgrey}{\textbf{adaptation strategies}} displayed (except QLoRA is not applied to GPT-3.5 and GPT-4). We evaluate each model's performance by comparing its outputs with expert clinician summaries. Each model paired with the adaptation strategy is evaluated using \textcolor{customgreen}{\textbf{quantitative similarity metrics}}. Finally, we perform a \textcolor{customyellow}{\textbf{clinical study}} where five clinicians rate three summaries (randomized order) for every summarization task: best-performing open-source model, best-performing proprietary model, and clinician-written.
The $^{*}$ indicates GPT-4's maximum context length at the time of experimentation (later increased to 128,000).}
\label{fig:pipeline}
\end{figure*}

We categorize the LLMs used in this study as open-source LLMs and proprietary LLMs as shown in Figure \ref{fig:pipeline}.

\subsubsection*{Open-Source LLMs}\label{subsubsec2}
The models in this category were trained either with a sequence-to-sequence (seq2seq) or autoregressive objective. The seq2seq architecture maps input sequences directly to output sequences. In contrast, autoregressive architectures predict the next token given the preceding context. The following models were considered:

\begin{itemize}
\item Clinical-T5-Large is a "text-to-text transfer transformer" \cite{lehman2023clinical} pre-trained on medical text, that employs transfer learning with the Sequence-to-Sequence (seq2seq) architecture. Clinical-T5-Large supports a context window of only 512 tokens.
\item FLAN-UL2 hails from the T5 \cite{raffel2020exploring} family, utilizing the seq2seq architecture. It employs a modified pre-training procedure to incorporate a context length of 2,048 tokens instead of only 512 tokens context length of the original FLAN-T5. FLAN-UL2 is further instruction tuned \cite{lampinen2022can} to enhance the model’s capability to understand complex narratives.
\item Llama2-13B stems from the Llama family of LLMs \cite{touvron2023llama} and is an open-source autoregressive model tailored for expanded pretraining on 2 trillion tokens. Llama2-13B allows up to 4,096 tokens as input.
\end{itemize}

\subsubsection*{Proprietary LLMs}\label{subsubsec3}
This category includes large-scale proprietary autoregressive models that utilize reinforcement learning with human feedback (RLHF) to further improve performance over instruction tuning using feedback from expert human evaluators. 
\begin{itemize}
\item GPT-3.5 (Azure API version GPT-3.5-Turbo-0613) \cite{openai2022chatgpt} contains 175 billion parameters and has been extensively fine-tuned for general tasks using human feedback, enhancing its ability to capture intricate details within any summarization task. This GPT-3.5 version can support input context lengths of up to 16,384 tokens.
\item GPT-4 (Azure API version GPT-4-0613) \cite{openai2023gpt4} achieves SOTA NLP performance and supports a context window of 32,768 (later expanded to 128,000) tokens, the highest compared to models listed in Figure \ref{fig:pipeline}, making it a suitable choice in multi-document summarization scenarios.
\end{itemize}

Clinical-T5-Large \cite{lehman2023clinical}, and FLAN-UL2 \cite{tay2022unifying} have a maximum input context length limit of 512 and 2,048 tokens, respectively. Hence, for the context length analysis in our results section, we only select models that allow at least 4,096 context length inputs, allowing exploration of model behavior for varying extents of the input length.

\subsection*{Adaptation Strategies}\label{sec5}
We integrate \textit{knowledge creation} \cite{albers2020assimilation} by applying a series of lightweight domain adaptation methods to pre-trained models as shown in Figure \ref{fig:pipeline}. The adaptation methods mentioned below gradually increase the level of adaptation to a downstream task.

\begin{itemize}
\item Null Prompting: This is a discrete zero-shot adaptation strategy \cite{wei2022chain} where the clinical note is supplied with a basic prompt, such as "brief hospital course". This is a baseline technique to evaluate how models respond to minimal guidance.
\item Prefix Prompting: Building upon the null prompting approach, we now provide a detailed instructional prompt \cite{lampinen2022can, wei2022chain}, like "summarize the clinical note," which serves to provide the model with context for the BHC synthesis task.
\item In-Context Prompting: We employ in-context learning (ICL) \cite{lampinen2022can} using a discrete few-shot prompt. ICL involves providing the LLM with examples within the prompt itself, demonstrating a desired behavior \cite{lampinen2022can}. We enhance our null prompting strategy by prepending one example clinical note and BHC pair, selected randomly from the training dataset.
\item Quantized Low-Rank Adaptation (QLoRA) \cite{hu2021lora, dettmers2024qlora}: This technique utilizes 4-bit quantization to enable parameter efficient fine-tuning \cite{ding2023parameter} by injecting rank-decomposition matrices into each layer of the model. We use QLoRA to fine-tune only open-source LLMs on the supervised task of BHC synthesis on a mix of NVIDIA A10G and V100 GPU machines. While model weights for proprietary LLMs are not publicly accessible, fine-tuning is possible through their API. However, technical details regarding the fine-tuning process remain private, making comparison with QLoRA fine-tuning difficult.

\end{itemize}

\subsection*{Experimental Setup}\label{sec6}

\subsubsection*{Quantitative Evaluation}

To identify the best-performing open-source and proprietary models, we conduct a performance analysis by randomly selecting 70 independent samples from the 100 test samples in the 0-1,024 context range. We reserve the remaining 30 independent samples for qualitative evaluation by clinicians. The 70:30 split between quantitative and qualitative analysis was determined to ensure a significant portion of the dataset is used for quantitative evaluation while keeping the workload for clinicians manageable. This separation also mitigates data leakage, providing a more objective assessment of model generalization, as the samples used for quantitative evaluation were not included in the reader study. To accurately assess the performance of models, we employ quantitative metrics commonly used in summarization tasks to evaluate the syntactic similarity and semantic relevance of the summaries. Bilingual Evaluation Understudy (BLEU) \cite{papineni2002bleu} evaluates the overlap between the generated and reference summaries using a weighted average of 1 to 4-gram precision. ROUGE-L \cite{lin2004rouge} assesses the precision and recall of the longest common subsequence. We focus only on ROUGE-L instead of other ROUGE variants to avoid redundancy, as it effectively captures essential n-gram overlap in our context and correlates well with other ROUGE variants \cite{liu2009exploring}. BLEU and ROUGE-L depict lexical overlap between generated and reference summaries. In contrast, we also use semantic similarity metrics like BERT-Score \cite{zhang2019bertscore} which leverages contextual BERT embeddings to evaluate the similarity between the generated and reference summaries. Finally, we perform an independent context-length analysis using fine-tuned Llama2-13B and GPT-4 with ICL, the best-performing LLMs allowing context lengths greater than 4,000 tokens as input using 100 test samples from each of the three context-length test sets from Table \ref{table:dataset_sample}b. Output lengths were set to the models' default configurations and were not specifically controlled in this analysis. To assess potential health equity implications, we perform a stratified subgroup analysis across patient-reported sex when assessing the quantitative performance of BHC summarization techniques. 

\subsubsection*{Qualitative Evaluation}

To evaluate the viability of \textit{clinical deployment} \cite{albers2020assimilation}, we conduct a study with five board-certified clinicians to compare BHCs written by clinicians during clinical practice to BHCs generated by our best-performing open-source and proprietary models using 30 clinical note-BHC pairs. We pair each clinical note with three BHCs: (1) original BHC written by an expert clinician, (2) BHC generated by the best-performing open-source LLM, and (3) BHC generated by the best-performing proprietary LLM. We present these pairs to five diverse clinicians from different institutions and backgrounds to reflect diversity in clinical practice. The clinicians each represent a clinical focus in internal medicine, nursing, thoracic radiology, pediatric radiology, and neuroradiology (with 11, 7, 5, 6, and 7 years of experience, respectively). The five clinicians rank the 30 note-summary pairs in randomized and blinded order on a Likert scale of 1 (poor) to 5 (excellent) based on four evaluation criteria:
\begin{itemize}
\item Comprehensiveness: How well does the summary capture important information? This assesses the recall of clinically significant details from the input text.
\item Conciseness: How well does the summary exclude non-important information? This compares how condensed, considering the value of a summary decreases with superfluous information.
\item Factual Correctness: How well does the summary agree with the facts outlined in the clinical note? This evaluates the precision of the information provided.
\item Fluency: How well does the summary exhibit fluency? This assesses the readability and natural flow of the content.
\end{itemize}

For the reader study, we gather diverse and well-informed feedback by ensuring that all participating clinicians possess at least five years of experience and a solid understanding of discharge summaries—a standard yet crucial task for clinicians. To reduce potential bias, we randomize the order of the three summaries, labeling them as 'Summary 1,' 'Summary 2,' or 'Summary 3.' All clinicians see questions in the same order, helping maintain consistency and reduce potential biases. To standardize score values, we provide a detailed rating scale across each aspect. For example, in rating comprehensiveness, a score of 1 indicates minimal capture of critical information, whereas a score of 5 denotes full coverage of key details. Furthermore, we guide clinicians to comparatively assess summaries within each note, ensuring that the scores consistently reflect differences in quality. To quantitatively evaluate the consistency of ratings, we calculate Krippendorff's alpha based on all scores provided by clinicians.

In total, we received 150 reviews across all five clinicians (30 clinical note-summary pairs for each of the five clinicians). As part of the study, we request optional qualitative feedback from clinicians on each note-BHC pair to dive deeper into the reasons behind clinician preferences. Of the 150 reviews, 37 received optional free-text comments from clinicians, while the others were left blank. We analyze all 37 comments to determine insights from observed common themes: 1) Factual Mistakes: Does the summary inaccurately represent critical information from the original note?, 2) Missing Critical Information: Does the summary omit critical information part of the original note?, 3) Hallucinations: Does the summary contain information that cannot be inferred from the original note?

\subsection*{Statistical Analysis}

We compare the best-performing proprietary and open-source LLMs with clinician summaries by exploring the following two hypotheses across each evaluation criterion: 1) Does our pool of five diverse clinicians exhibit a preference for LLM-generated summaries over clinician-written ones?, and 2) Does our pool of five diverse clinicians exhibit a preference for summaries generated by adapted proprietary LLMs over summaries generated by adapted open-source LLMs?. We use the non-parametric Wilcoxon signed-rank tests with a significance level $\alpha$ $=0.05$, and with Bonferroni corrections using XLSTAT.

\begin{figure}[!t]%
\centering
\includegraphics[width=1\linewidth]{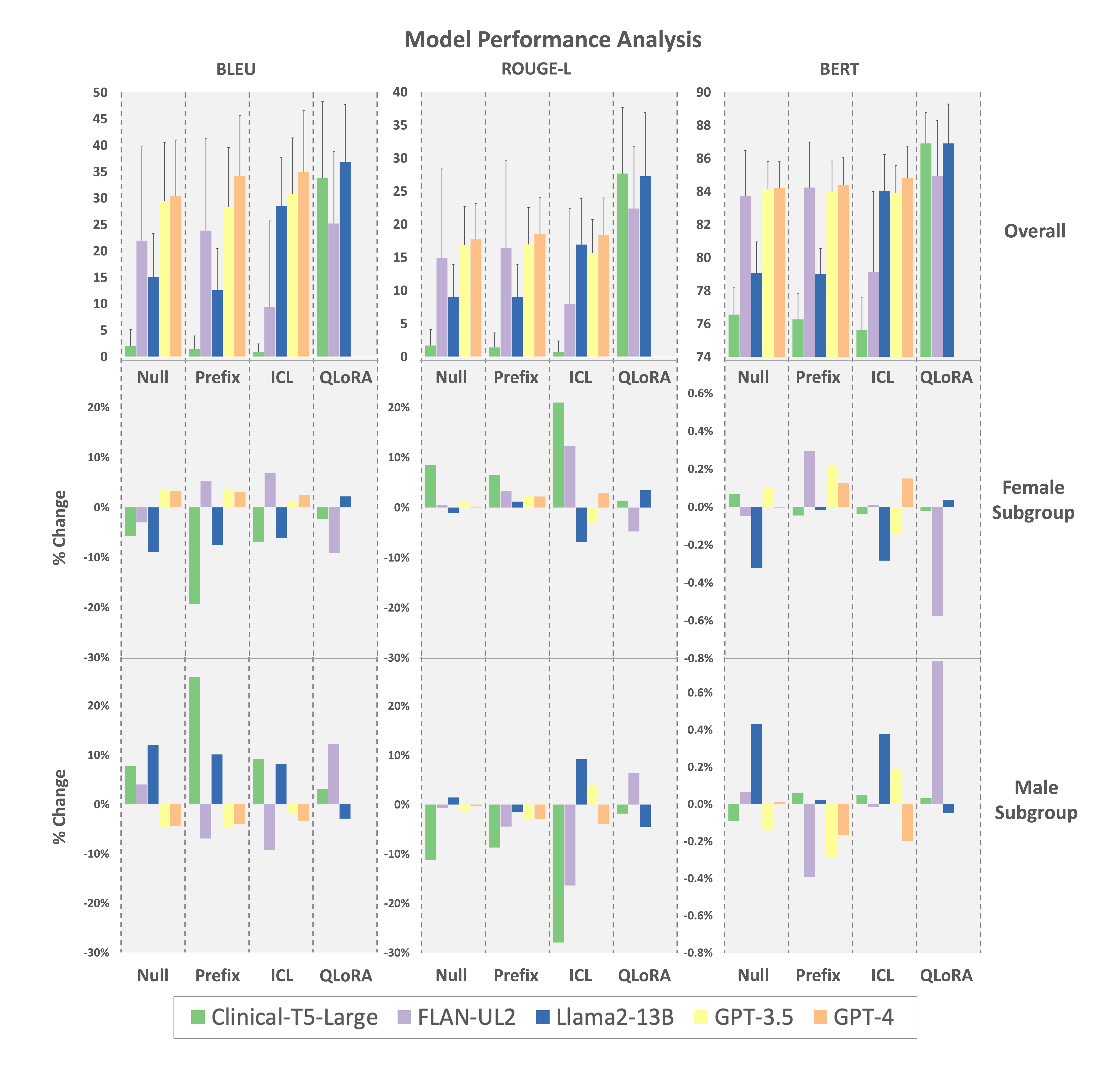}
\caption{Quantitative metric results for each choice of model, across increasing domain-adaptation strategies. In summary, QLoRA as an adaptation strategy outperforms other adaptation methods. Specifically, QLoRA Llama2-13B outperforms other models in BLEU score, while achieving comparable performance to Clinical-T5-Large in BERT-Score and ROUGE-L.}
\label{fig:quant_metrics}
\end{figure}

\section*{RESULTS}

\subsection*{Model Performance Analysis}

Despite being the smallest model, Clinical-T5-Large (750M parameters), exhibits competitive performance after QLoRA fine-tuning and achieves the largest performance improvement with increasing adaptation (Figure \ref{fig:quant_metrics}). FLAN-UL2 displays strong performance across all adaptation strategies, except ICL. Its improvement in performance with fine-tuning is limited. Llama2-13B's performance improves substantially following QLoRA fine-tuning, achieving the highest BLEU and BERT scores across all LLMs and strategies. GPT-3.5 shows good performance but exhibits limited benefits with increasing adaptation. GPT-4 with ICL outperforms GPT-3.5, becoming the top-performing proprietary LLM. When evaluating the deviation of each model's performance across the subgroups of patient-reported sex, we notice large variations with the Clinical-T5-Large model (Figure \ref{fig:quant_metrics}). With increasing adaptation, GPT models show higher variance from their original ROUGE-L and BERT scores. LLMs after QLoRA adaptation generally show lower variations among subgroups than other adaptation strategies.

\begin{figure}[!t]%
\centering
\includegraphics[width=1\linewidth]{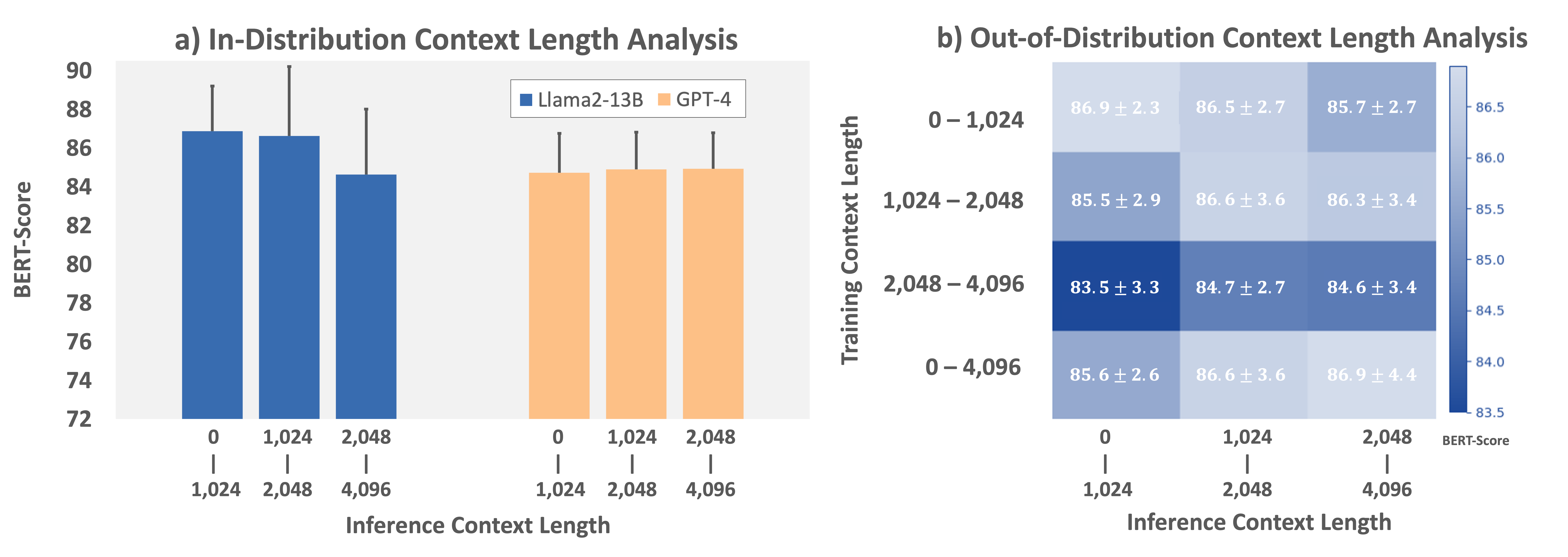}
\caption{\textbf{a)} Quantitative evaluation metrics across increasing input context lengths. GPT-4 shows consistency in performance whereas Llama2-13B shows a drop in summarization with increasing context length inputs. \textbf{b)} Context size analysis for QLoRA Llama2-13B (in/out-of-distribution), where each item on the y-axis displays an independent model fine-tuned on samples from a specific context length range. The summarization performance of the combined model trained on 0 - 4,096 context length inputs outperforms other models with longer input clinical notes at inference (more than 1,024 tokens).}
\label{fig:in-dist}
\end{figure}

\subsection*{Context Length Analysis}

We compare the performance of LLMs when the adaptation (training) and inference (testing) context lengths are identical (in-distribution) (Figure \ref{fig:in-dist}a). QLoRA Llama2-13B exhibits a drop in in-distribution performance. In contrast, in-context GPT-4 displays consistent performance in-distribution as the context length of the train and test set increases. Overall, in-context GPT-4 exhibits more robustness than QLoRA Llama2-13b with increasing context lengths. We explore LLM performance with differing training and testing context lengths (Figure \ref{fig:in-dist}b). We observe that models at testing perform best in-distribution. The performance of models trained on smaller context-length samples deteriorates when testing on longer-context-length samples.

\begin{figure}[t]
\begin{center}
\includegraphics[width=1\linewidth]{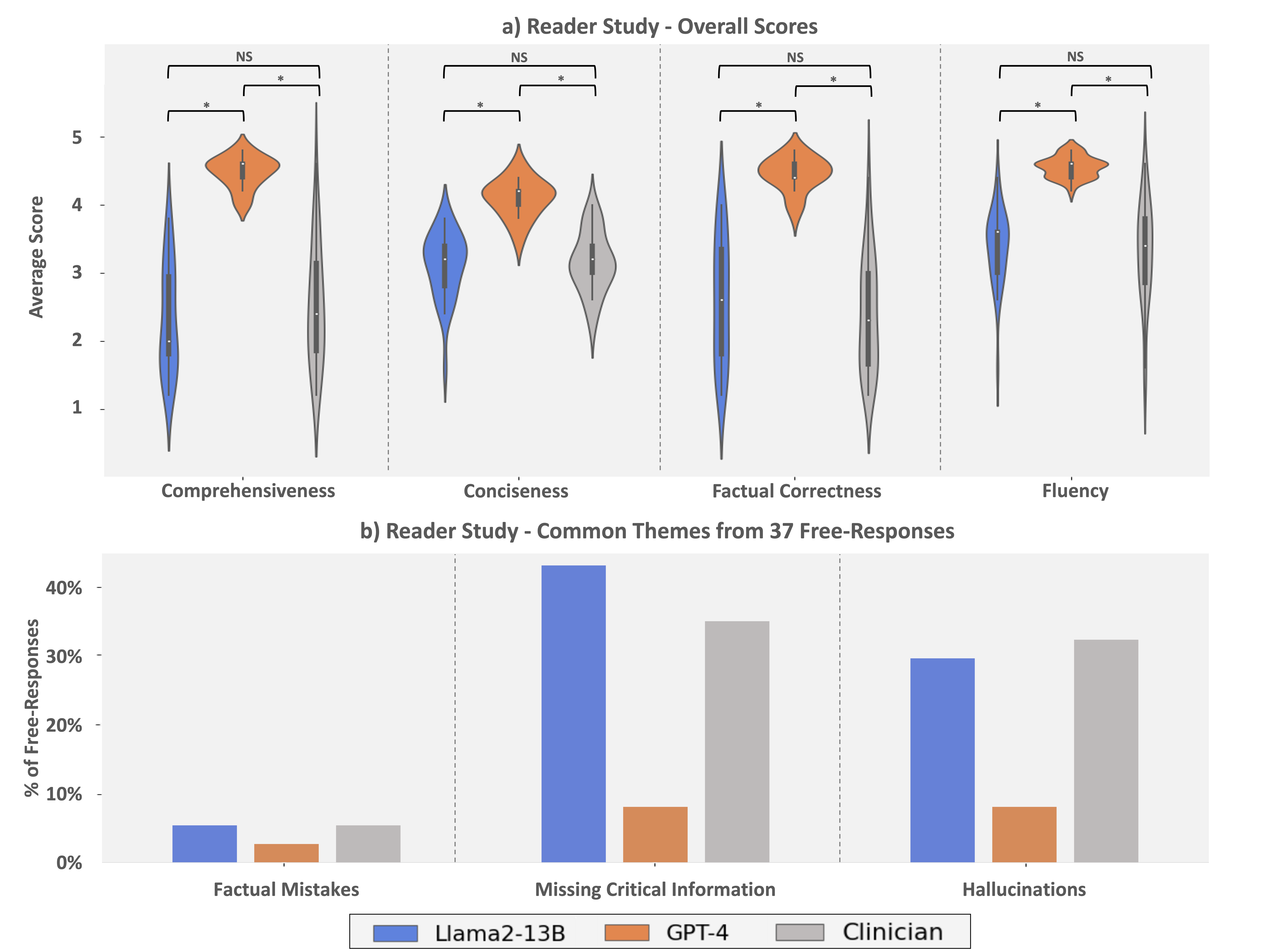}
\end{center}
\caption{\textbf{a)} Violin plot showing results from the reader study with five clinicians. Clinicians exhibit a strong preference for in-context GPT-4 (adapted large-scale proprietary LLM) summaries over QLoRA Llama2-13B (adapted open-source LLM) and clinician-written summaries with statistical significance by the Wilcoxon signed-rank test (*$p < 0.001$) across each attribute. (NS: Not Significant). \textbf{b)} Plot showing common themes derived from detailed reader comments. This sub-analysis reiterates the preference for in-context GPT-4 while exhibiting comparable performance of open-source models and clinicians.}
\label{fig:reader}
\end{figure}

\subsection*{Clinical Reader Study}
\label{subsec:clinical-reader-study}
We find that our five clinical readers strongly prefer in-context GPT-4 summaries for attributes of comprehensiveness, conciseness, factual correctness, and fluency, compared to the original clinician-written summaries ($p < 0.001$) and QLoRA Llama2-13B summaries ($p < 0.001$) (Figure \ref{fig:reader}a). However, we find no significant differences between reader preferences comparing Llama2 summaries and the original clinician-written summaries ($p = 0.05$). Overall, GPT-4 scores have a narrower range, showcasing lower variation. We report the proportion of scores from readers between the range of 3 to 5 for GPT-4, Llama2-13B, and clinician summaries: 100\%, 47\%, and 40\%, respectively. To further explore reasons behind clinician preferences, we analyze 37 optional "free-text" comments provided by clinicians and report common themes with respective frequencies in Figure \ref{fig:reader}b. In-context GPT-4 summaries consistently report the least amount of cases across all themes: factual mistakes, omission of critical information, and hallucinations. QLoRA Llama2-13B and clinician-written summaries exhibit similar frequencies across all themes. As part of the reader study, we observe moderate consistency among raters given the inherently subjective nature of the task (Krippendorff's $\alpha = 0.47$). Overall, this study underscores the strength of LLMs in generating summaries that are both statistically superior and preferred by our panel of clinicians in the majority of cases.

\section*{DISCUSSION}
Recent adaptations of LLMs in the clinical domain emphasize their potential in understanding medical language. Noteworthy recent works \cite{van2023radadapt, van2024clinical, chen2022toward, lyu2023translating, patel2023chatgpt, singh2023chatgpt} use open-source datasets and display the performance of LLMs in multiple summarization tasks in medicine: radiology reports, patient health questions, progress notes, doctor-patient conversations, and discharge summaries. Other works \cite{wang2024drg, singhal2023large} further explore domain adaptation of LLMs in the clinical domain through fine-tuning for improved performance in medical tasks. Our benchmark shows that open-source LLMs can produce high-quality BHCs and that Llama2-13B comes out as a superior choice for further analysis of open-source LLMs based on its performance and an expanded context length window (Figure \ref{fig:quant_metrics}a). The subgroup analysis (Figure \ref{fig:quant_metrics}b) further emphasizes the importance of a multi-faceted analysis when evaluating models. Furthermore, our context-length analysis (Figure \ref{fig:in-dist}) showcases the benefit of fine-tuning LLMs on datasets with a wide range of context lengths. A deeper qualitative analysis through our reader study (Figure \ref{fig:reader}) suggests that adapted proprietary LLMs like GPT-4 can outperform other adapted LLMs as well as clinicians, while open-source LLMs like Llama2-13B can produce summaries similar in quality to clinician summaries. However, open-source models provide the benefit of ease in implementation, where sending data via an API may not be possible due to privacy constraints. Our benchmark also reports strong summarization performance from Clinical-T5-Large, while it is the smallest parameter model, possibly due to extensive pre-training on medical text. However, Clinical-T5-Large also presents a large variation in performance across subgroups, making it less suitable.

For our quantitative benchmarking study, we employ a targeted evaluation strategy by including a slew of metrics whose individual strengths cover the weaknesses of other metrics. We reported ROUGE-L, BERT-Score, and BLEU for shorter context length pairs (0-1,024 tokens) as these metrics remain widely used for evaluating summaries in this length regime. For longer context length pairs (>1,024 tokens), we report only BERT-Score, recognizing the potential shortcomings of ROUGE-L for longer summaries. ROUGE-L primarily captures syntactic matches rather than semantic similarity, which can result in limited content coverage and a preference for summaries that closely follow the original wording \cite{lin2004rouge, zhang2019bertscore, koh2022far, van2024clinical}. BERT-Score offers a more semantic similarity assessment, though it too may underrepresent fine-grained information overlap, particularly in long-form texts \cite{zhang2019bertscore}. We also acknowledge that there is currently no perfect metric for evaluating complex, long-form summarizations in clinical settings \cite{van2024clinical}.

Our study bridges computational and clinical disciplines by combining quantitative NLP metrics with qualitative clinical assessments, ensuring the evaluation reflects both technical performance and clinical relevance. It is crucial to recognize that quantitative measures, such as NLP metrics, and qualitative assessments by clinical readers gauge different aspects of summarization quality. NLP metrics quantify the level of (textual, semantic, contextual) similarity between two summaries, whereas clinical readers inform perceptual (subjective) quality differences. We observe that fine-tuned Llama2-13B consistently outperforms in-context GPT-4 in quantitative similarity metrics. However, the reader study results present a nuanced narrative where clinicians express a preference for GPT-4-generated summaries. This finding reflects a critical point: high quantitative similarity scores do not necessarily translate to a "better summary" from a clinician’s perspective, as well as across subgroups. Enhancing the alignment between quantitative metrics and qualitative assessments will enable a more holistic evaluation for clinical text summarization \cite{van2024clinical}.

While categorizing models as strictly open-source or proprietary may not comprehensively capture the many differences between the models, we opt for this categorization because fine-tuning proprietary models is possible but only with unreported mechanisms, whereas, fine-tuning open-source LLMs can lead to repeatable and verifiable models. For the broader quantitative study, we justify the model choice by aiming for diverse models, considering factors like model accessibility and architecture. In contrast, for the qualitative study, we select the best model from each of the open-source and proprietary categories based on their performance in the quantitative study. From the proprietary LLM category, we select GPT-4 (SOTA at the time of experimentation), serving as a robust reference for proprietary models, and GPT-3.5, a smaller, more cost-effective comparison within the same category. Due to resource constraints and the proprietary nature of fine-tuning mechanisms for GPT-3.5 and GPT-4, we limit fine-tuning to open-source models with up to 20 billion parameters (such as Llama2-13B, Clinical-T5-Large, and FLAN-UL2), as these can be fine-tuned on consumer-grade GPUs.

While noting the strengths of our study, we also discuss its limitations. The first limitation we observe is that because Clinical-T5-Large can only accept inputs of less than 512 tokens, the quality of the Clinical-T5-Large generated summaries in our experiments might be impacted. However, recent works \cite{van2023radadapt} have demonstrated the strong performance of domain-adapted seq2seq models from the T5 family in radiology report summarization, motivating our inclusion of this model. Moreover, based on prior work \cite{fleming2024medalign}, given strong decoder models, there is potential to use multi-stage refinement and summarization techniques to condense the size of the input prompt itself. Another limitation of our work is the limited availability of publicly available datasets for clinical note summarization, making it difficult to gauge the performance of our adapted LLMs in diverse scenarios across hospitals and clinical practices. The MIMIC-IV dataset is derived from a specific U.S.-based healthcare system, and its generalizability to other healthcare systems worldwide may be limited. Similarly, the clinician population in our study may not be fully representative of broader clinical populations. Additional public clinical note datasets may help alleviate these challenges. We also acknowledge that our clinician reader mix, primarily composed of radiology subspecialists with one clinical physician, may affect the generalizability of findings, as discharge summaries are typically reviewed by clinical doctors. A final limitation of our work is that newer adaptation strategies and LLMs, such as Llama-3 (released after our experimentation phase), could not be incorporated into this study. However, our framework, and dataset are designed to be model agnostic, enabling future studies to evaluate newer LLM families and adaptation strategies as they emerge. We encourage further research to iterate on these advancements using the MIMIC-IV-BHC dataset \cite{aali2024mimic}.

\section*{CONCLUSION}
In this investigation, we develop the MIMIC-IV-BHC \cite{aali2024mimic} dataset and perform a comprehensive quantitative and qualitative evaluation of using LLMs for synthesizing BHCs using adaptation strategies for both open-sourced and closed-sourced models. Via a clinical reader study, we depict that adapted open-source models can match the quality of clinician-written summaries, while adapted proprietary models can outperform the quality of clinician-written summaries across dimensions of comprehensiveness, conciseness, factual correctness, and fluency. This suggests LLMs could streamline documentation, reduce errors, and enhance clinical workflows, improving patient safety. Overall, this study provides a framework for evaluating LLMs, showcasing their potential for reducing clinicians’ documentation burden and improving patient outcomes through better documentation practices.

\clearpage
\section*{CONTRIBUTORSHIP STATEMENT}
AA designed and executed the entire study, led the acquisition and development of the novel dataset, and drafted the manuscript. DVV provided essential methodological expertise, contributed to experimental design, and assisted in data analysis and manuscript writing. YIA contributed to the clinical study analysis and assisted in manuscript preparation. JH, CB, EPR, SG, and NC provided clinical expertise by reviewing clinical notes with their respective summaries and contributing to the manuscript review. JD, AST, and JK contributed to dataset development, LLM pipeline establishment, and provided critical feedback on the manuscript. ASC oversaw the project and contributed to study design, data interpretation, and manuscript writing. All authors contributed to a critical review of the manuscript and approved the final version.

\section*{FUNDING STATEMENT}
A.C. receives research support from NIH grants R01 HL167974, R01 HL169345, R01 AR077604, R01 EB002524, R01 AR079431, and P41 EB027060; from NIH contracts 75N92020C00008 and 75N92020C00021; from Stanford Center for Artificial Intelligence and Medicine, Stanford Institute for Human Centered AI, from Stanford Center for Digital Health, from Stanford Cardiovascular Institute, from Stanford Center for Precision Health and Integrated Diagnostics; from GE Healthcare, Philips and Amazon. C.B. received research support independent of this project from the ProMedica Foundation, Chur, Switzerland. Computing resources were partially provided by One Medical and Stanford University. Microsoft provided Azure OpenAI credits for this project via the Accelerate Foundation Models Academic Research (AFMAR) program. 

\section*{DATA AVAILABILITY STATEMENT}
To allow researchers to replicate our adaptation and evaluation approach for BHC summarization, we publicly release:
\begin{enumerate}
    \item The MIMIC-IV-BHC dataset, published on PhysioNet \cite{aali2024mimic}.
    \item The underlying code for data pre-processing, LLM adaptation, and evaluation: \href{https://github.com/StanfordMIMI/clin-bhc-summ.git}{github.com/StanfordMIMI/clin-bhc-summ}.
\end{enumerate}

For this study, we use Microsoft Azure API versions GPT-3.5-Turbo-0613 and GPT-4-0613, as permitted under the guidelines set by PhysioNet, following all dataset usage protocols. Furthermore, this study was institutional review board exempt because it did not meet the criteria for human subjects research. Due to the use of MIMIC data in fine-tuning, we are unable to publicly share the fine-tuned model weights. However, these weights are available to credentialed users of MIMIC datasets upon request.

\section*{COMPETING INTERESTS STATEMENT}
All authors declare no financial or non-financial competing interests. 

\clearpage
\printbibliography
\end{document}